\documentclass{article}
\pdfoutput=1

\usepackage{arxiv}
\usepackage[export]{adjustbox}
\usepackage[para,online,flushleft]{threeparttable}
\usepackage{booktabs}
\usepackage{threeparttable}
\usepackage{makecell}
\usepackage[utf8]{inputenc} 
\usepackage[T1]{fontenc}    
\usepackage{hyperref}       
\usepackage{url}            
\usepackage{booktabs}       
\usepackage{amsfonts}       
\usepackage{nicefrac}       
\usepackage{microtype}      
\usepackage{graphicx}
\usepackage[caption=false]{subfig}
\usepackage{doi}
\usepackage{amsmath}
\usepackage{multirow}
\usepackage{multicol}
\usepackage{titlesec}
\usepackage[utf8]{inputenc}
\usepackage[T1]{fontenc}
\usepackage{lineno}
\usepackage{xcolor}
\usepackage{float}
\titlespacing*{\section}{0pt}{0.3\baselineskip}{0.3\baselineskip}
\titlespacing*{\paragraph}{0pt}{0.3\baselineskip}{0.3\baselineskip}

\title{DeepPatent2: A Large-Scale Benchmarking Corpus for Technical Drawing Understanding}

\author{Kehinde Ajayi\\
	Old Dominion University\\
	Norfolk, VA 23529 \\
	\texttt{kajay001@odu.edu} \\
    \And
    Xin Wei\\
    Old Dominion University\\
    Norfolk, VA 23529 \\
    \texttt{xwei001@odu.edu} \\
   \And
    Martin Gryder\\
   Old Dominion University\\
    Norfolk, VA 23529 \\
    \texttt{mgryd001@odu.edu} \\
	\And
    Winston Shields\\
    Old Dominion University\\
    Norfolk, VA 23529 \\
    \texttt{wshie002@odu.edu} \\
   \And
    Jian Wu\\
	Old Dominion University\\
	Norfolk, VA 23529 \\
	\texttt{j1wu@odu.edu} \\
   \And
    Shawn M. Jones\\
    Los Alamos National Laboratory\\
    Los Alamos, NM 87545 \\
    \texttt{smjones@lanl.gov} \\
     \And
    Michal Kucer\\
    Los Alamos National Laboratory\\
    Los Alamos, NM 87545 \\
    \texttt{michal@lanl.gov} \\
     \And
   Diane Oyen\\
    Los Alamos National Laboratory\\
    Los Alamos, NM 87545 \\
    \texttt{doyen@lanl.gov} \\
}

\renewcommand{\shorttitle}{DeepPatent2: A Large-Scale Benchmarking Corpus for Technical Drawing Understanding}

\begin{document}
\maketitle

\begin{abstract}
Recent advances in computer vision (CV) and natural language processing have been driven by exploiting big data on practical applications. However, these research fields are still limited by the sheer volume, versatility, and diversity of the available datasets. CV tasks, such as image captioning, which has primarily been carried out on natural images, still struggle to produce accurate and meaningful captions on sketched images often included in scientific and technical documents. The advancement of other tasks such as 3D reconstruction from 2D images requires larger datasets with multiple viewpoints. We introduce DeepPatent2, a large-scale dataset, providing more than 2.7 million technical drawings with 132,890 object names and 22,394 viewpoints extracted from 14 years of US design patent documents. We demonstrate the usefulness of DeepPatent2 with conceptual captioning. 
We further provide the potential usefulness of our dataset to facilitate other research areas such as 3D image reconstruction and image retrieval.

\end{abstract}


\section*{Background \& Summary}
Technical illustrations, sketches, and drawings are images constructed to convey information more straightforwardly to humans than using text alone \cite{carney2002pictorial,mayer2019illustrations,sketchy}. One goal of computer vision is to build models to understand information contained in these images. Specific tasks include recognizing objects and determining their attributes, capturing the relationships between sub-images, and understanding the context of objects \cite{Nadeem2019}. Natural images such as those contained in MS COCO \cite{lin2014microsoft} and ImageNet \cite{imagenet} have been used to develop solutions to these tasks using deep neural networks. Different from natural images \cite{lin2014microsoft,imagenet}, technical drawings are a type of image frequently found in design patents. Although they usually do not contain many features of natural images such as various colors, gradient, and environmental detail, drawings often abstract away unnecessary distractions leaving the strokes, lines, and shading that are sufficiently detailed so that drawn objects and aspects are still recognizable by humans. Compared with natural images, technical drawings, are under-studied by the computer vision and information retrieval communities. 

Recently, several sketch-based datasets have been developed, e.g., \cite{vrochidis2012concept,opensketch}. \emph{QuickDraw} consists of 50M images sketched by letting users draw designated objects within a short time \cite{QuickDraw50M}. Other sketch datasets are much smaller, including TU-Berlin \cite{eitz2012humans}, Sketchy \cite{sketchy}, and sheepMarket \cite{sheepMarket}. These free-hand sketches tend to have very few strokes and limited viewpoints. Therefore, they are not suitable for tasks that rely on drawing details, like understanding scientific and technical information. CLEF-IP 2011 provides two patent image datasets of 10k images of heterogeneous types, including flow charts and chemical structures, used for patent retrieval and classification into 9 classes \cite{piroi2011clef}. \emph{ImageNet-Sketch} contains 50k images of 1000 classes, based on search results of text queries against Google Image \cite{wang2019imagenetsketch}. 
The technical drawing data we introduce here is different from these datasets in that it contains enriched semantic information, including object names and multiple views of the same object. Our dataset also contains {segmented} figures. Compared with \emph{QuickDraw}, our dataset contains more details for drawings, diverse object names (more than 132,000), and viewpoint information (Table~\ref{tab:comparing}). Recently, Kucer et al. \cite{Kucer_2022_WACV} introduced {\sc DeepPatent}, a dataset of over 350K images from design patents for image retrieval. However, the {\sc DeepPatent} dataset does not include identification of the objects or descriptions of their viewpoints. In addition, compound technical drawings were not segmented to isolate individual figures.

\begin{table}[H]
    \centering
    \begin{tabular}{@{}ll@{}rcc@{}}
    \toprule
     \textbf{Dataset} & \textbf{Size} (number of images) & \textbf{\#Categories} & \textbf{Captions} & \textbf{Viewpoints}\\ \midrule
     ImageNet-1k \cite{russakovsky2015imagenet1k} & 2M photos  & 1,000 & No & No\\  
     ImageNet-21k \cite{imagenet} & 14M photos  & 21,000 & No & No\\ 
     WebVision \cite{li2017webvision} & 2.4M photos & 1,000 & No & No \\
     Flickr-8K \cite{flickr8k} & 8K photos & - & Yes & No\\ 
     Flickr-30K \cite{young2014image} & 30K photos & - & Yes & No\\ 
     \midrule
     ImageNet-Sketch \cite{wang2019imagenetsketch} & 50K drawings & 1,000 & No & No \\
     TU-Berlin \cite{eitz2012humans} & 20K sketches & 250 & No & No\\
     QuickDraw \cite{QuickDraw50M} & 50M+ sketches & 345 & No & No\\
     Sketchy \cite{sketchy} & 75K sketches, 12K photos & 125 & No & No\\
     Sheep 10K \cite{sheepMarket} & 10K sheep sketches & 1 & No & No\\
     Sketch Flickr15K \cite{hu2013performance} & 330 sketches, 15K photos & 33 & No & No \\ 
     CLEF-IP 2011 \cite{piroi2011clef} & 10k diagrams & 9 & No & No \\
     DeepPatent \cite{Kucer_2022_WACV} & 350K+ drawings & - & No & No\\
     \midrule
     DeepPatent2 & 2M+ drawings & 132,890 & Yes & Yes\\ 
     \bottomrule
    \end{tabular}
    \caption{{\sc DeepPatent2} compared with major sketch-based and natural image datasets. \#Categories means the number of object names. }  
    \label{tab:comparing}
\end{table}

\begin{figure}[H]
    \centering
    \includegraphics[width=0.95\textwidth]{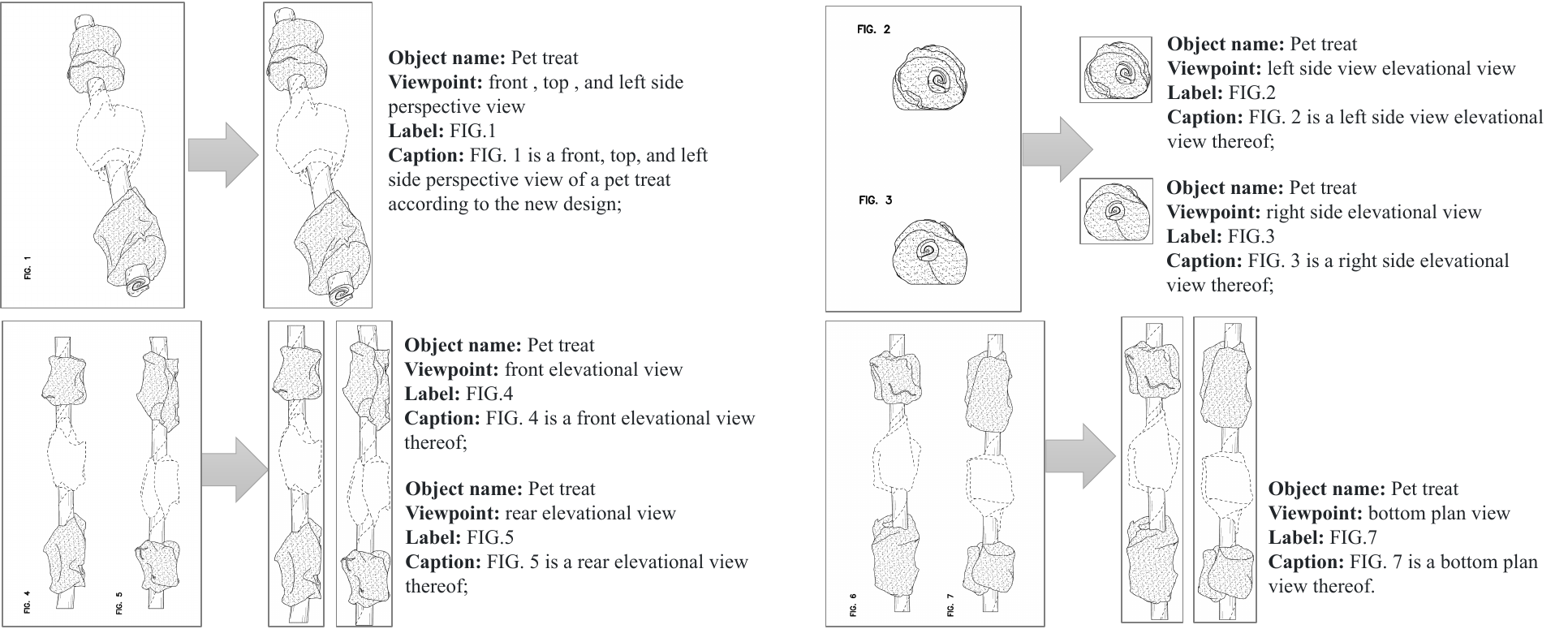}
    \caption{An example from {\sc DeepPatent2} extraction results for US Design Patent \#0836880.}
    \label{fig:example}
\end{figure}

In this paper, we introduce a new dataset, {\sc DeepPatent2}, consisting of more than 2 million automatically segmented and tagged technical drawings from more than 300,000 design patents granted from 2007 to 2020. Figure~\ref{fig:example} shows an example of the figures segmented and tagged from a single patent. We use an approach reminiscent of a large-scale image collection for real-world images with real human tags; similar to WebVision \cite{li2017webvision}, which consisted of 2.4 million images crawled from the Web with associated metadata. It was demonstrated that the noisy web images like WebVision were sufficient for training a good deep learning model for visual recognition \cite{li2017webvision}. We propose a novel pipeline using natural language processing (NLP) models to extract object names and viewpoints from figure captions, using computer vision (CV) methods to segment compound figures, and aligning text information with corresponding figures. The pipeline is developed based on our solid results on label extraction \cite{gong2021recognizing}, visual descriptor extraction \cite{wei2022jcdl}, and image segmentation \cite{hoque2022aaai}. To demonstrate the usefulness of our dataset, we train baseline deep-learning models on conceptual captioning and demonstrate that the performance of the models benefits significantly from an increasing size of training data. We expect that {\sc DeepPatent2} will facilitate training robust neural models on other tasks such as 3D image reconstruction and image retrieval for technical drawings. 

\section*{Methods: Building the DeepPatent2 Dataset}
\label{sec:methods}
\subsection*{Overview}
{\sc DeepPatent2} contains over 2 million technical drawings from the United States Patent and Trademark Office USPTO \cite{USPTO} design patent documents published from 2007 to 2020.  Our dataset extends the {\sc DeepPatent} dataset \cite{Kucer_2022_WACV} in three aspects.
\begin{enumerate}
    \item {\sc DeepPatent2} is more than {5 times} larger than {\sc DeepPatent};
    \item {\sc DeepPatent2} contains both \emph{original} and \emph{segmented} patent drawings;
    \item The metadata of each drawing contains object name and viewpoint information, automatically extracted using a supervised sequence-tagging model with high accuracy. 
\end{enumerate}  

\begin{figure}[H]
    \centering
    \includegraphics[width=\textwidth]{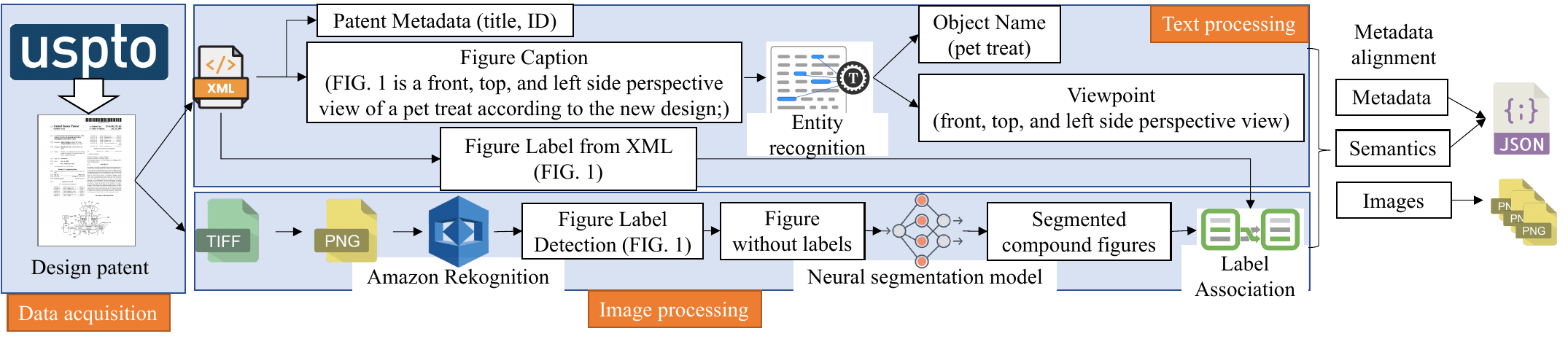}
    \caption{The architecture of the pipeline to create {\sc DeepPatent2}. Text in parentheses are examples. The matching module links semantic information parsed from captions to individual images segmented from original compound figures.}
    \label{fig:arch}
\end{figure}

The pipeline to create the dataset is illustrated in Figure~\ref{fig:arch}. The pipeline includes three major components, namely, data acquisition, text processing, and image processing. A patent document includes an XML file containing textual content and TIFF files containing associated images. One patent TIFF file may contain several patent figures, as shown in Figure~\ref{fig:example}, which we refer to as a compound figure. In the following, we refer to a ``figure file'' as an image file containing a single or a compound figure. We refer to a ``figure'' as an individual figure (e.g., a figure with a unique label such as ``FIG.~1"). 

The text processing step aims at automatically tagging individual figures with text that a person would use to describe them. In addition to extracting the design category information included in the patent XML file, 
we extract human-readable object names from figure captions. One challenge to this goal is that although the XML document contains captions and inline references for individual figures, the document does not directly map figures to figure files, because a figure file may contain multiple figures. For example, Figure~\ref{fig:example} contains 7 individual figures in 4 figure files. To overcome this challenge, we develop modules that first segment compound figures, resolve figure labels, and link them to text processing results to associate captions to their respective figures. The final data includes JSON files containing metadata, automatically extracted descriptions (object names and viewpoints), and images of individual figures, as well as original figure files. 

In the following subsections, we elaborate on each module in the pipeline. All computation was performed on a Dell server with 4 NVIDIA GTX 2080 Ti GPUs, 24 hyperthreaded Intel Xeon Silver 4116, 300GB RAM, and 7TB disk space. In addition, we applied AWS Rekognition's DetectText service on all patent figure files. The total size of the dataset is approximately 380GB before compression. 

According to USPTO, design patent drawings are illustrations of a manufactured object's design, which includes detailed information about contours, shapes, material texture, properties, and proportions. Drawings and text of patent documents are public domain; and are created by writers, artists, and inventors with the knowledge that these works will become public domain \cite{uspto-tos}.

\subsection*{Patent Data Acquisition}
\begin{figure}[H]
    \centering
    \includegraphics[width=0.8\textwidth]{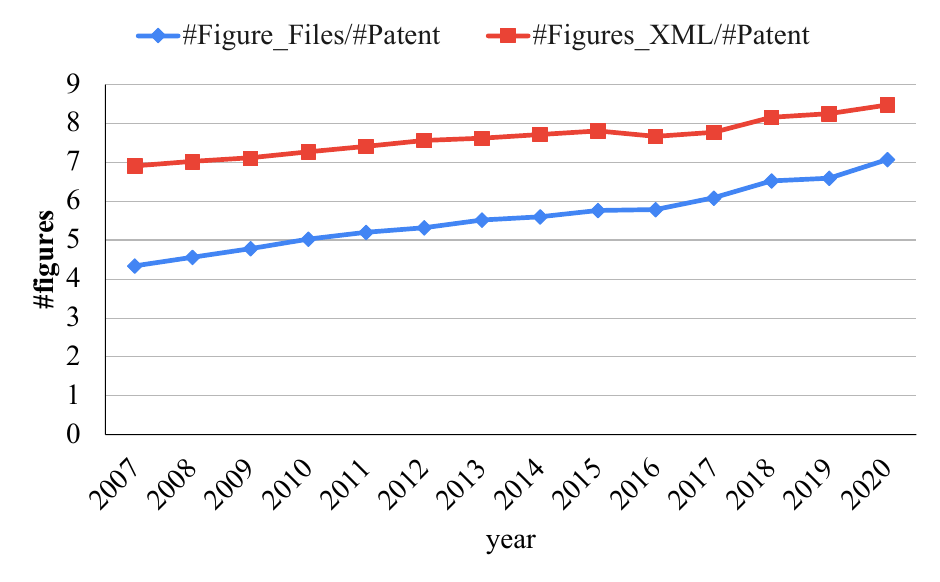}
    \caption{Average numbers of figures and subfigures per patent each year from 2007 to 2020.}
    \label{fig:trend}
\end{figure}

We collected patent data ranging from the years 2007 to 2020 from the USPTO website in the form of zip and tar files. These files consist of both the full-text (in XML format) and figures (in TIFF format). Each TIFF file contains one or multiple figures. Table~\ref{tab:count} shows statistics of the raw data. XML files before 2006 had a different schema and did not include separate XML tags for individual figures and subfigures, which may introduce further errors to figure label parsing and label-figure alignment, so we only include patents after 2007. The discrepancy between figure numbers parsed from XML (\#Figure XML) and figure numbers from segmentation (\#Figure Segmented) reflects the imperfection of the text and image processing methods, which will be incorporated in assessing the data quality. Figure~\ref{fig:trend} shows the average number of figure files per patent and the average number of figures per patent of our dataset, demonstrating the trends of the increasing popularity of using figures in patent applications over the last 14 years. 

\subsection*{Text Processing: Entity Recognition}
\label{sec:text_processing}
\begin{figure}[H]
    \centering
    \fbox{\includegraphics[width=0.9\textwidth]{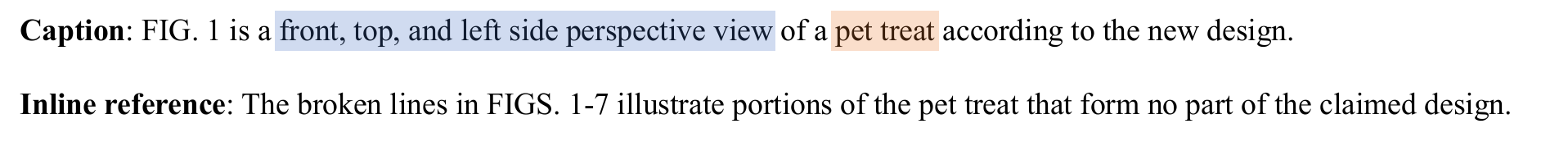}}
    \caption{An example of a figure caption and its corresponding inline reference in a patent XML file. Orange highlighted text stands for  object names, and blue highlighted text stands for viewpoints.}
    \label{fig:caption}
\end{figure}

The text processing pipeline includes parsing plain text from XML documents and extracting semantic information from figure captions. In patent documents, figure captions are usually enclosed by special XML tags, so they can be accurately extracted. An inline reference is a sentence in the body text that cites one or multiple figures. We used regular expressions to match figure tags (e.g., ``FIG.'', ``FIGS.'', etc.) and then extracted complete sentences as inline references. Figure~\ref{fig:caption} shows an example of a figure caption and its corresponding inline reference in an XML file. Each individual figure has a caption, which we use to extract object names and viewpoints. 

The detail of the text processing pipeline is elaborated in Wei et al. (2022) \cite{wei2022jcdl}. Here, we highlight the best model and its performance. We treated this task as an entity recognition problem. The text was first tokenized. Each token was encoded as a feature vector by a pre-trained word embedding model. Word vectors were then fed to a sequence-tagging neural network. We compared the BiLSTM-CRF (bidirectional long short-term memory or BiLSTM followed by a conditional random field or CRF) with a transformer model. Both were usually believed as the state-of-the-art sequence-tagging models \cite{li2022survey}. We compared several word embedding models, including GloVe \cite{pennington2014glove}, RoBERTa fine-tuned on GPT-2 \cite{radford2019language}, the original RoBERTa \cite{liu2019roberta}, BERT \cite{devlin2019bert}, ALBERT \cite{lan2020albert}, and DistilBERT \cite{sanh2019distilbert}. The sequence-tagging model incorporated context into the initial feature vector of each word. The CRF layer classified each token under the IOB schema. The best performance was achieved using DistilBERT \cite{sanh2019distilbert} with the BiLSTM-CRF architecture. The average F1-measures for the overall entity recognition, object name, and viewpoint were $0.960$, $0.927$, and $0.992$, respectively. 

The ground truth corpus was created by randomly selecting 3300 patent figure captions from the US design patents. Each caption is manually annotated by two researchers independently using {\sc brat}, a web-based text annotation tool \cite{stenetorp2012brat}. An example of an annotated caption is shown in Figure~\ref{fig:caption}. The ground truth was split into training, validation, and test sets, each consisting of 2700, 300, and 300 captions respectively. We then applied this model to all patent figure captions and extracted a total of {132,890 unique object names and 22,394 unique viewpoints}. The result of post-extraction validation using a random sample of 100 figure captions was consistent with the evaluation based on the benchmark. 

\subsection*{Image Processing: Figure Segmentation and Metadata Alignment}
The goal of the image processing pipeline is segmenting compound figures and recognizing figure labels, which is necessary because the patent documents do not contain information that maps individual figures' captions to figure files. This process has four steps (Figure~\ref{fig:arch}). (1) Figure label detection, including text and positions; (2) Segmenting compound figure files into individual figures; (3) Associating labels with individual figures; (4) Aligning metadata with individual figures.  The final dataset product is publicly available at the Harvard Dataverse repository \cite{DVN/UG4SBD_2023}.

\setlength{\tabcolsep}{4pt}
\begin{table}[h]
\centering
  \begin{tabular}{lrrr}
    \toprule
    {\bf OCR Tools}  &   \textbf{Recall} & \textbf{Precision} & \textbf{F1}   \\
     \midrule
     AWS Rekognition with rectifier  &  \textbf{96.03}   &  \textbf{97.58}  &   \textbf{96.80}  \\
     AWS Rekognition without rectifier &  79.37   & 81.97  &   80.65  \\
     \midrule
     Google Vision API  &  88.90   &  100.00  &   94.10  \\
    AWS Textract & 80.16 & 99.05  &  88.61 \\
 Tesseract \cite{smith2007overview}  &  44.40   &  96.60  &   60.84  \\
 EAST deep net trained on ICDAR2015 \cite{zhou2017east,karatzas2015icdar} & 11.90 &  14.29  &   12.99  \\
    \bottomrule
  \end{tabular}
   \caption{A comparison of OCR engines on extracting labels and bounding boxes in patent drawings.}
  \label{tab:performance}
\end{table}

\subsubsection*{Figure Label Detection}
\label{sec:recognizing_fig_labels}
The original figure files are in TIFF format, which is not compatible with many optical character recognition (OCR) engines and computer vision packages. Therefore, we converted TIFF to PNG format, which is widely used in computer vision and does not introduce compression artifacts as in JPEG files. We use AWS Rekognition's DetectText service, a commercial OCR engine, to recognize figure labels, including text content and bounding boxes \cite{vijayarani2015performance}. 

To evaluate the quality of extraction results, we extended our previous work \cite{gong2021recognizing} and compared several top-performing OCR engines using a corpus consisting of 100 randomly selected design patent figures from the USPTO dataset in 2020 including both single and compound figures. The evaluation metrics included precision, recall, and F1. The precision was calculated as the number of correctly recognized labels divided by the total number of labels captured. The recall was calculated as the number of correctly recognized labels divided by the total number of labels shown on sampled figures. The results (Table~\ref{tab:performance}) indicate that AWS Rekognition achieves the highest F1 ($96.80\%$),  compared with other OCR engines. In particular, Rekognition achieves the highest recall ($96.03\%$). Google Vision API shows an excellent F1 ($94.10\%$), but the recall score is relatively low ($88.90\%$). Tesseract, an open-source OCR engine, achieves a relatively high precision ($96.60\%$) but a poor recall ($44.40\%$). Therefore, we adopted Rekognition into our pipeline.

The errors made by Rekognition are due to several reasons. Errors may happen when the labels are rotated by 90 degrees, resulting in reversed order of label tokens. For example, ``Fig.3'' in a single figure could be  recognized as [``3'', ``Fig.''], and [``FIG.7'', ``FIG.6''] in a compound figure could be recognized as [``7'', ``6'', ``FIG.'', ``FIG.'']. This is likely because Rekognition scans the image from top to bottom.
In addition, unlike many other OCR engines that output a single result, Rekognition outputs multiple results scored from 0 to 1. To address the two issues above, we developed a script that rectifies the Rekognition output when the confidence score output by Rekognition is above a threshold $\theta_0=0.5$ and then reconstructs the labels by ``gluing'' tokens. This rectifier improved the F1 score of figure label recognition by $16\%$, from $0.806$ to $0.968$ (Table~\ref{tab:performance}). 

The results of this step are figure labels and their bounding boxes for each figure file. The figure label regions in the images are whitened out before figure files are passed to the segmentation module. 

\subsubsection*{Compound Figure Segmentation}\label{sec:segmentation}
A significant number of US patents use compound figures, containing more than one individual figure, so they need to be segmented before they are associated with captions. We proposed a transfer learning method, with a transformer-based neural model called MedT \cite{dosovitskiy2021medt} pre-trained on compound medical images and then fine-tuned on a small set of annotated patent figures. We compared our model against several state-of-the-art image segmentation models on a test set containing annotated patent figures. We then associated segmented figures with labels using a proximity-based method. The details of our approach can be found in Hoque et al. (2022) \cite{hoque2022aaai}. Here, we outline the pipeline and highlight the performance of the best models. 

To our best knowledge, there was no labeled dataset that could be used for patent figure segmentation, therefore we developed an in-house ground truth dataset consisting of 500 figure files randomly selected  from design patents. Here, we assume the number of figures in each figure file equals the number of labels detected by the OCR engine. If the two numbers do not equal, the segmentation is flagged as an error. The results will still be included in the final dataset as challenging cases for future models, but these cases can be easily excluded using the flags. In our ground truth of 500 figures, there were 480 compound figures each containing up to 12 individual figures. Each figure has been annotated by a graduate student using the VGG Image Annotator \cite{dutta2019via} by drawing bounding boxes around individual figures. We performed an independent human validation of bounding boxes. 

As a baseline, we proposed an unsupervised method called \emph{Point-shooting} \cite{hoque2022aaai}. The idea is to randomly ``shoot'' open dots onto a compound figure. Dots that overlay any strokes are filled with a single solid color and the other dots are removed, so the remaining dots created a region filled with single color pixels. A contour can be drawn around the region, outlining the profile of individual figures. A bounding box can then be drawn around the contour. Using this method, we can obtain a mask of individual figures and their bounding boxes. The bounding boxes can be directly used for comparing against the ground truth. Point-shooting achieves an accuracy of 92.5\%, calculated as the percentage of individual figures that are correctly segmented. 

\emph{MedT} was designed to train a transformer-based semantic segmentation framework \cite{dosovitskiy2021medt}. The core component is a gated position-sensitive axial attention mechanism, designed to overcome the limitation of a vanilla transformer so that a more robust model can be trained using relatively small training sets. Because the point-shooting method was able to generate masks of individual figures with relatively high accuracy, we used the output of the point-shooting method as the training data for MedT. As our result would show, the MedT model achieved higher performance on the noisy training data. The training figures were first passed through a convolution block before passing through a global branch, which captures dependencies between pixels and the entire image. The same figure is broken down into patches, which were passed through a similar convolutional block before passing through a local branch, which captures dependencies among neighboring pixels. A re-sampler aggregates the outputs from the local branch and generates the output feature maps. The outputs from both branches are finally aggregated followed by a $1\times1$ convolutional layer to pool these feature maps into a segmentation mask. We fine-tuned the pre-trained MedT model on our training set and achieved an accuracy of 97\% with a much shorter runtime ($\approx1/35$) on the test set compared with the point-shooting method. MedT outperformed point-shooting and other deep learning baselines including U-Net \cite{ronneberger2015u}, HR-Net \cite{wang2020hrnet}, and DETR \cite{carion2020end}.

\subsubsection*{Label Association and Metadata Alignment}\label{sec:la}
We have two sources of metadata: labels recognized by the OCR engine and semantic information parsed from XML files. To generate the contextualized figures, two alignments were conducted. 

\paragraph{Label Association} This step associates labels output by the OCR engine with segmented figures. This was treated as a bipartite matching problem. A general solution is the Ford-Fulkerson method \cite{ford1956bpm}. Because of the specialty of our problem, we used a simple heuristic method by matching an individual figure segmented from a compound figure with the closest label recognized by the OCR engine. The proximity was calculated as the Euclidean distance between the geometric centers between the bounding boxes of labels and figures. Despite the simplicity of the method, it achieves an accuracy of {97\%}, evaluated on a set of {200} randomly selected figures in the matching results. The errors in the label-figure alignment are mainly attributed to upstream errors made by the OCR engine and the segmentation model, which occurs in {$\approx 7.5\%$} figures ({\#Figure Mismatch} / {\#Figure Segmented}) (Table~\ref{tab:count}). Here, (\#Figure Mismatch) refers to the number of figures that could not be aligned with labels using the proximity method above. Another type of error is attributed to a compact and irregular arrangement of labels. The output of this step is a set of individual figures with labels. We mark cases in which the number of labels recognized does not equal the number of segmented figures. For completeness, we still include these cases in the data product. Users can easily remove these cases in downstream training/analysis or use them to design better algorithms to automatically correct errors. 

\begin{table}[H]
        \centering
        \begin{tabular}{r|r|c|c|c|c|c}
        \toprule
            \textbf{Year}& \multicolumn{1}{c|}{\bf\#Patent} & \multicolumn{1}{c|}{\bf\#Figure File} & \multicolumn{1}{c|}{\bf\#Figure XML} & \multicolumn{1}{c|}{\bf\#Figure Segmented} & \multicolumn{1}{c|}{\bf\#Figure Mismatch} & \multicolumn{1}{c}{\textbf{Mismatch\%}}\\
            \midrule
            2020 &  34,895 & 246,971 & 295,693& 296,929& 18,417 & 6.2\% \\
            2019 &  34,813 & 229,557 & 287,122& 276,276 & 17,639 & 5.8\% \\
            2018 &  30,513 & 199,015 & 249,137 & 246,352 & 14,555 & 5.9\% \\
            2017 &  30,878 & 187,913 & 240,197 & 239,748 & 17,385 & 7.2\% \\
            2016 &  28,886 & 167,177 & 221,684 & 219,822 & 17,098 & 7.7\% \\
            2015 &  26,000 & 149,891 & 202,949 & 201,297 & 15,508 & 7.7\% \\
            2014 &  23,666 & 132,661 & 182,609 & 181,279 & 14,471 & 7.9\% \\
            2013 &  23,478 & 129,538 & 179,031 & 177,580 &  14,626 & 8.2\% \\
            2012 &  21,959 & 116,938 & 166,015 &  164,399 & 15,490 & 9.4\% \\
            2011 &  21,361 & 111,089 & 158,255 & 157,038 & 12,906 & 8.2\% \\
            2010 &  17,082 & 85,857 & 124,131 & 122,680 & 9,345 & 7.6\% \\
            2009 &  23,116 & 110,512 & 164,448 & 162,466 & 12,772 & 7.8\% \\
            2008 &  25,565 & 116,531 & 179,478 & 176,424 &  14,143 & 8.0\% \\
            2007 &  24,063 & 104,424 & 166,312 & 163,472 &  13,651 & 8.3\% \\
           \midrule
            \textbf{Total} & 366,275 & 2,088,074 & 2,817,031  & 2,785,762 & 208,006 &--\\
            \bottomrule
        \end{tabular}
        \caption{The number of design patents (\#Patent), figure files (\#Figure file), subfigures recorded in XML (\#Figure XML), segmented individual figures (\#Figure segmented), mismatched figures due to OCR errors and label-figure alignment errors (\#Figure mismatch), and the mismatch rate (Mismatch\%), calculated as (\#Figure mismatch)/(\#Figure segmented), i.e., Mismatch\% is defined as the number of mismatched segmented figures divided by the total number of segmented figures. 
        }
        \label{tab:count}
\end{table}

\paragraph{Metadata Alignment} This step aligns labeled figures with captions extracted from XML files by matching figure labels parsed from XML files and labels associated with individual figures in the last step. The result of this step contains segmented figures, each having  metadata, including the label, the caption, the bounding box coordinates, and document-level metadata, including patent ID, year, title, number of figures, etc. The method used here is based on strict integer matching, so the errors in the final data are caused by errors propagated from upstream processes. 

\begin{table}[ht]
    \centering
    \begin{tabular}{r|p{9cm}|p{5cm}}
    \toprule
    {\bf Field Name}     &  \multicolumn{1}{c|}{\bf Meaning} & \multicolumn{1}{c}{\bf Example}\\
    \midrule
    id     & A unique number representing an individual figure segmented from a compound figure & 1 \\ 
    \midrule
    patentID & The patent ID followed by the patent approval date &USD0836880-20190101 \\ 
    \midrule
    patentdate & The patent approval date &2019-01-01 \\ 
    \midrule
    figid & Segmented/individual figure label& 1 \\ 
    \midrule
    caption & The figure caption directly extracted from the patent XML file & FIG. 1 is a front, top,  and left side perspective view of a pet treat according to the new design; \\ 
    \midrule
    object & The object name automatically extracted from the figure caption& Pet treat \\  
    \midrule
    aspect & The viewpoint automatically extracted from the figure caption& front, top, and left side perspective view \\ 
    \midrule
    figure\_file & The original figure file name & USD0836880-20190101-D00001.png \\
    \midrule
    subfigure\_file & The segmented figure file name as shown & USD0836880-20190101-D00001\_1.png \\ 
    \midrule
    object\_title & The title of the patent & Pet treat \\ 
    \midrule
    x\_figure & The x coordinate of the upper left vertex of the segmented figure's bounding box in pixels, measured from the upper-left-corner of the original figure& 614 \\ 
    \midrule
    y\_figure & The y coordinate of the upper left vertex of the segmented figure's bounding box in pixels, measured from the upper-left-corner of the original figure &91 \\ 
    \midrule
    w\_figure& The width of the segmented figure's bounding box in pixels & 969 \\ 
    \midrule
    h\_figure & The height of the segmented figure's bounding box in pixels & 2740 \\ 
    \midrule
    W\_label & The width of the segmented figure's label's bounding box, normalized with respect to the original figure & 0.0439147 \\ 
    \midrule
    H\_label & The height of the segmented figure's label's bounding box, normalized with respect to the original figure &0.0502626 \\ 
    \midrule
    L\_label & The normalized coordinate of the upper left vertex of the segmented figure's bounding box, measured from the upper-left corner of the original figure &0.119197 \\ 
    \midrule
    T\_label & The normalized coordinate of the upper left vertex of the segmented figure's bounding box, measured from the upper-left corner of the original figure & 0.861215 \\ 
    \bottomrule
    \end{tabular}
    \caption{The fields of an individual figure in the JSON file in the final data.}
    \label{tab:jsonschema}
\end{table}

\section*{Data Records} 
\label{sec:datarecords}

\begin{figure}[ht]
    \centering
    \includegraphics[width=0.8\textwidth]{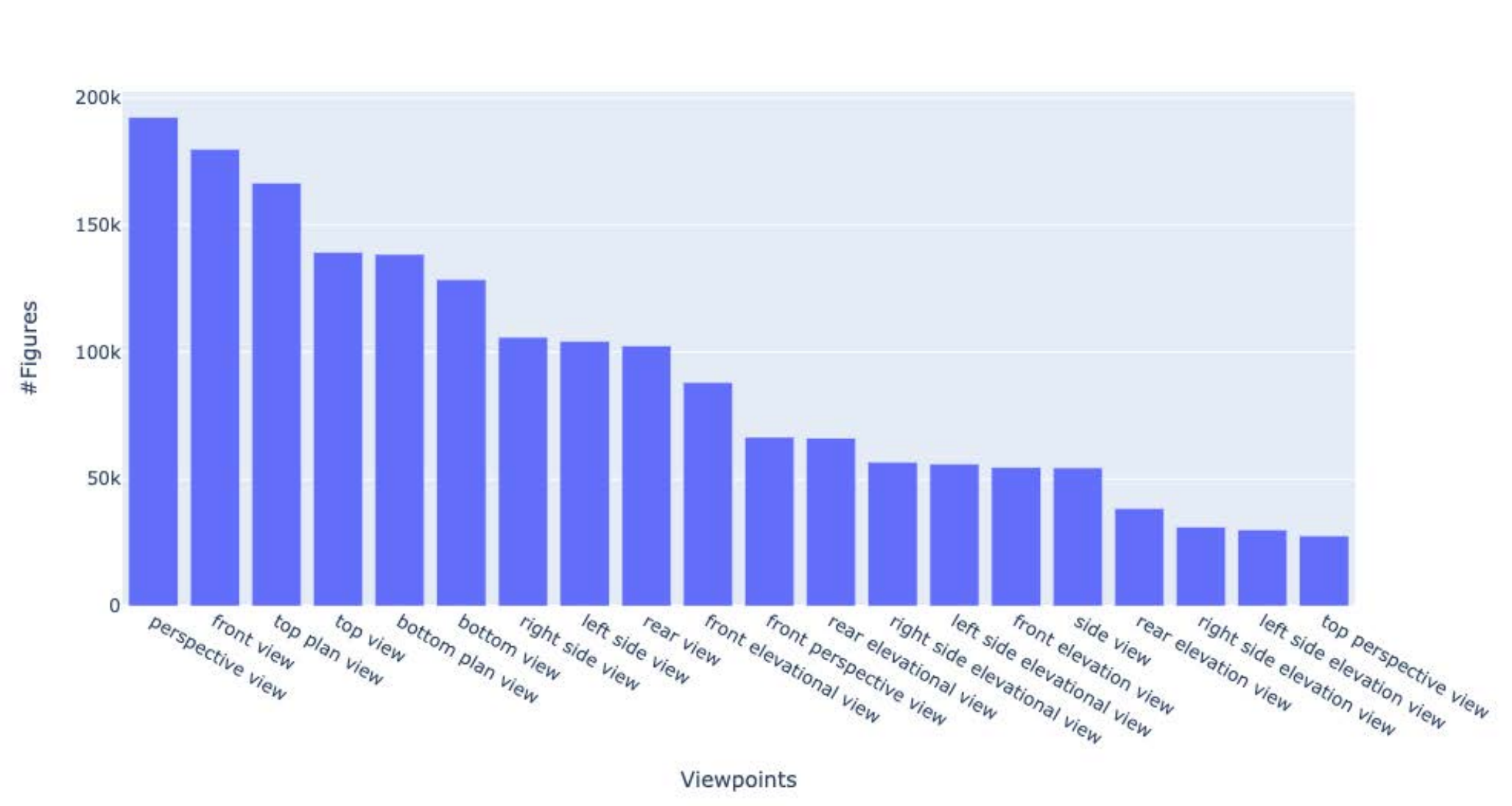}
    \caption{Distribution of viewpoints identified in DeepPatent2. Only the top 20 viewpoint are shown.}
    \label{fig:topviewpoint}
\end{figure}
The final data product contains about 2 million compound PNG figures, 2.7 million segmented PNG figures, and metadata in JSON format. The dataset is organized by year  (Table~\ref{tab:count}). The total size of the compressed dataset is 314 GB. The size of each year's data is roughly proportional to the number of figure files (Figure~\ref{fig:trend}). 

\begin{figure}[ht]
    \centering
    \includegraphics[width=.7\textwidth]{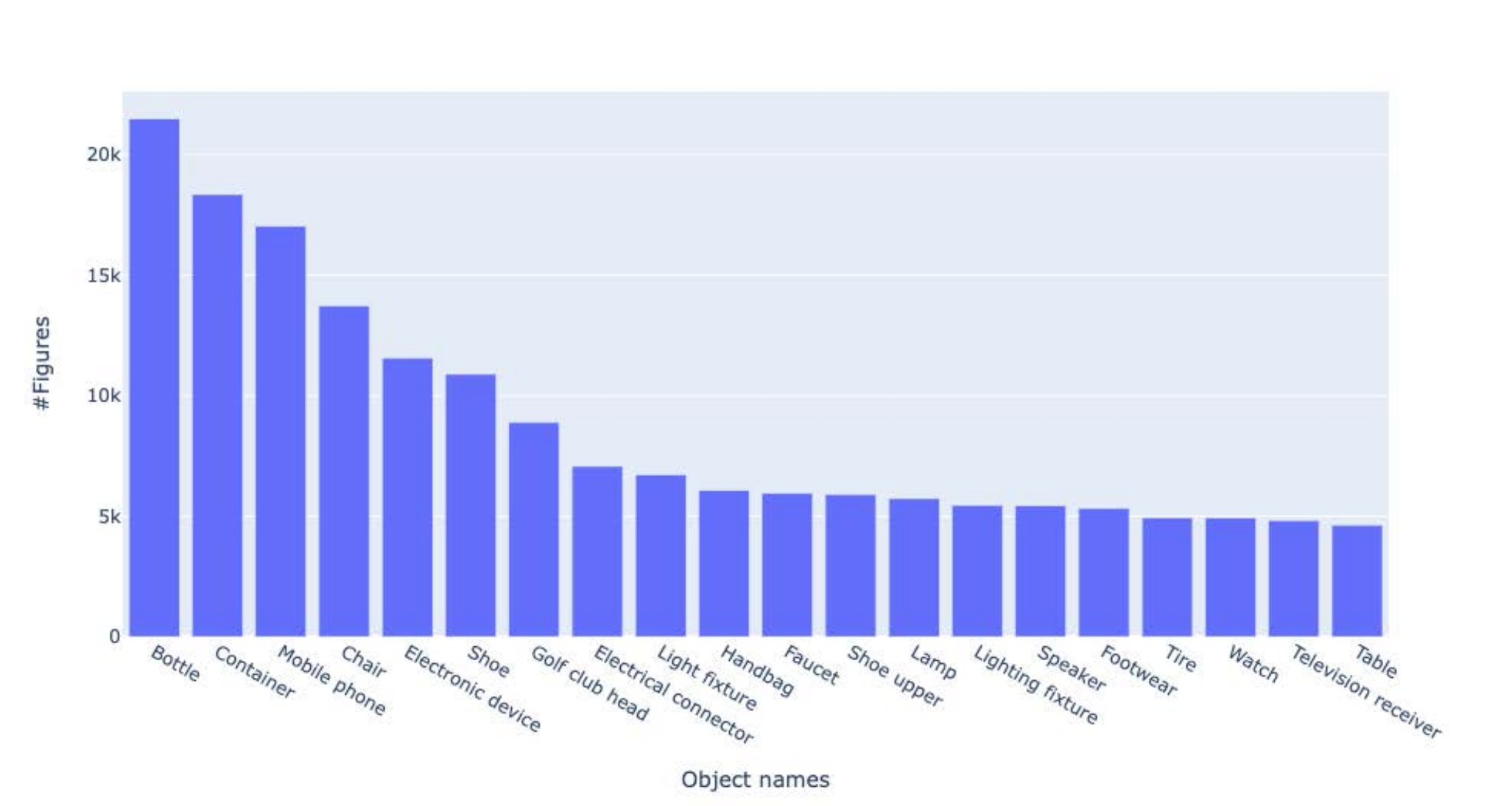}
    \caption{Distribution of object names identified in DeepPatent2. Only the top 20 object names are shown. }
    \label{fig:topobject}
\end{figure}

The JSON files contain document-level and figure-level metadata (one JSON file per year). Each entry in the JSON file represents a segmented figure, which is tagged with the patent ID, original figure file, the object name, and viewpoints extracted, the figure labels, the bounding boxes of segmented figures, and their labels. The schema and a sample record of this JSON file are shown in Table~\ref{tab:jsonschema}. The patent captions contain an average of 12 words with a vocabulary of about 25,000 tokens. The semantic information extraction resulted in a total of 132,890 unique object names and 22,394 viewpoints. The distribution of the top 20 object names is shown in Figure~\ref{fig:topobject}. The distribution of the top 20 viewpoints is shown in Figure~\ref{fig:topviewpoint}. 

\begin{figure}[ht]
    \centering
    \includegraphics[width=0.6\textwidth]{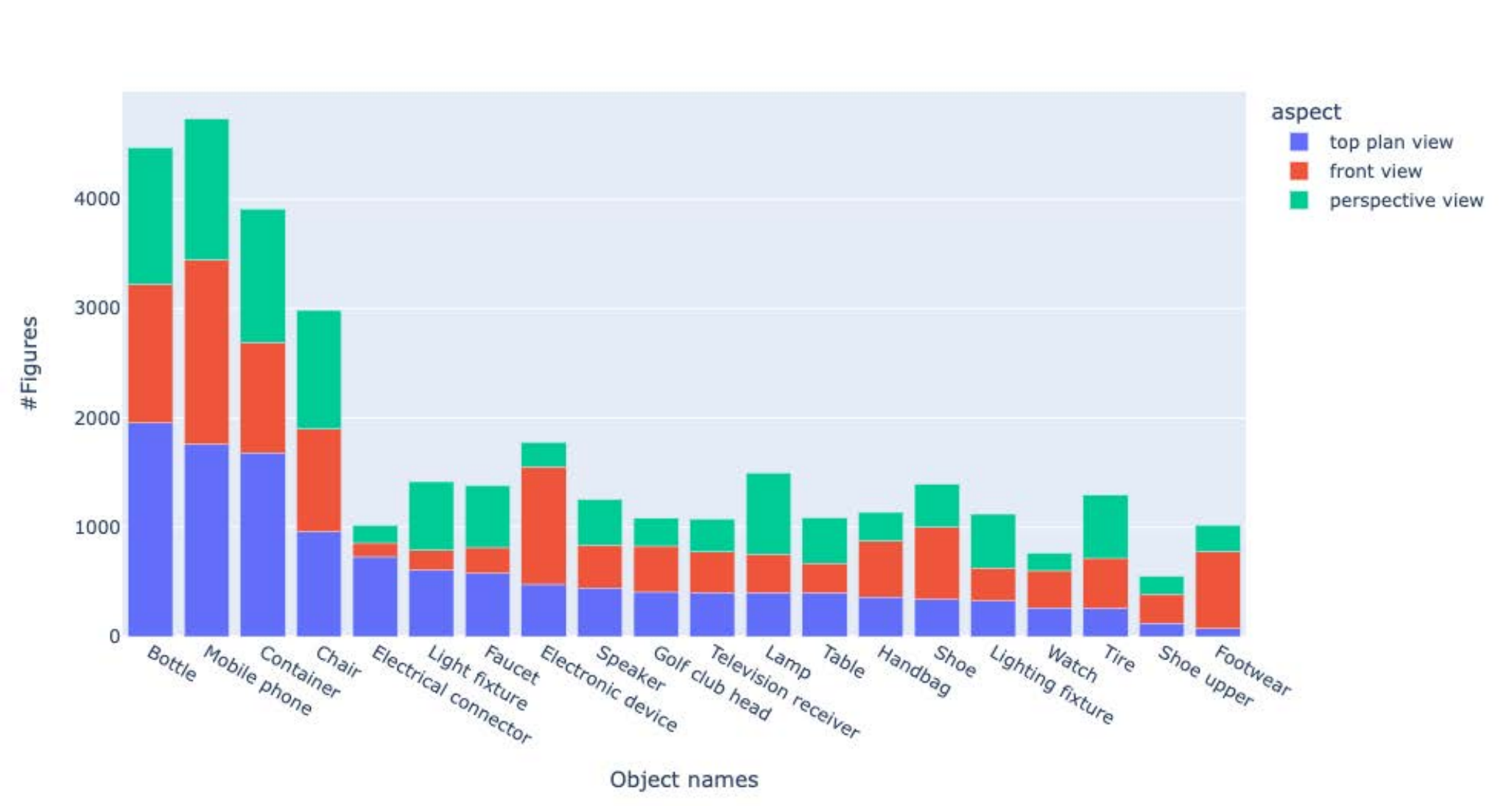}
    \caption{Distribution of top 20 objects with the top three viewpoints color-coded. Note that the bar heights are different from Figure~\ref{fig:topobject} because not all aspects were included. The object names along the abscissa are sorted by the total number of figures.}
    \label{fig:topobjectviewpoint}
\end{figure}

The \emph{perspective view}, \emph{front view}, and \emph{top plan view} are the top 3 viewpoints used in design patents. The object distribution color-coded by the top three viewpoints (Figure~\ref{fig:topobjectviewpoint}) indicates that objects are depicted with diverse and disproportionate viewpoints, which poses challenges for 3D reconstruction from 2D sketches. The complete distribution is provided with the dataset. 

\begin{figure}[ht]
    \centering
    \includegraphics[width=0.5\textwidth]{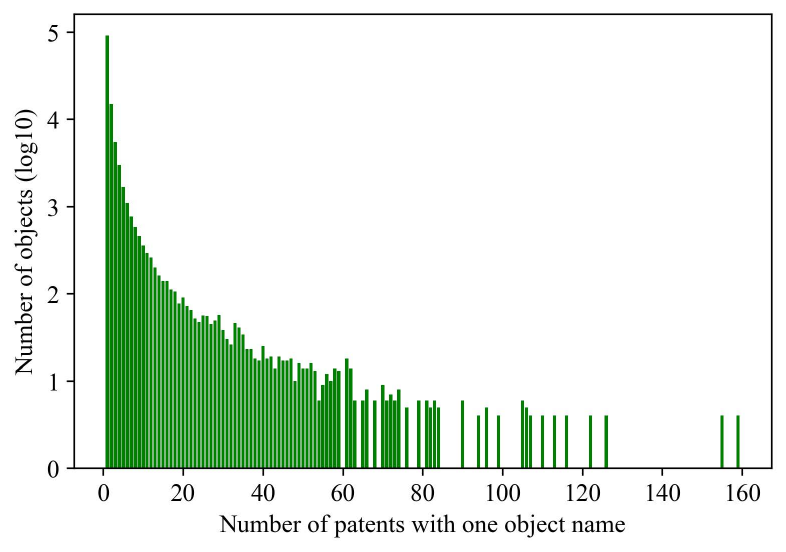}
    \caption{One object name may be described by multiple patents. This figure shows a truncated distribution of the number of patents in one object category. For example, when the $x$-axis value is 7, the $y$-axis value is 774, meaning each of the 774 objects is described by 7 patents. The $x$-axis is truncated because data points beyond the truncation point are sparse.}
    \label{distribution_patent}
\end{figure}

We observe that one object name, such as ``bottle'', may appear in multiple patents. 
Figure~\ref{distribution_patent} shows the distribution of the number of patents per object. Similarly, Figure~\ref{distribution_subfigure} shows the distribution of the number of individual figures per object. The peak of Figure~\ref{distribution_patent} happens when the $x$-axis value is 1, meaning that most frequently, there is 1 patent for each object name. The peak of Figure~\ref{distribution_subfigure} happens when the $x$-axis value is 7, meaning that most frequently, there are 7 individual figures for each object.

The USPTO design patents contain class labels that were manually assigned using the Locarno International Classification system, which was developed by members of the Paris Convention for the Protection of Industrial Property and administered by the International Bureau of the World Intellectual Property Office WIPO \cite{WIPO}. The system has 32 major categories and 16 minor categories. The fraction of patents assigned with minor categories is less than 0.5\%. A full Locarno classification code is composed of two pairs of numbers separated by a hyphen with the first pair showing the top-level design classes and the second pair showing the low-level classes. Locarno classification code is available in the first section of each patent document. They follow a fixed format and thus can be accurately parsed for all patents in our dataset. 

\begin{figure}[ht]
    \centering
    \includegraphics[width=0.7\textwidth]{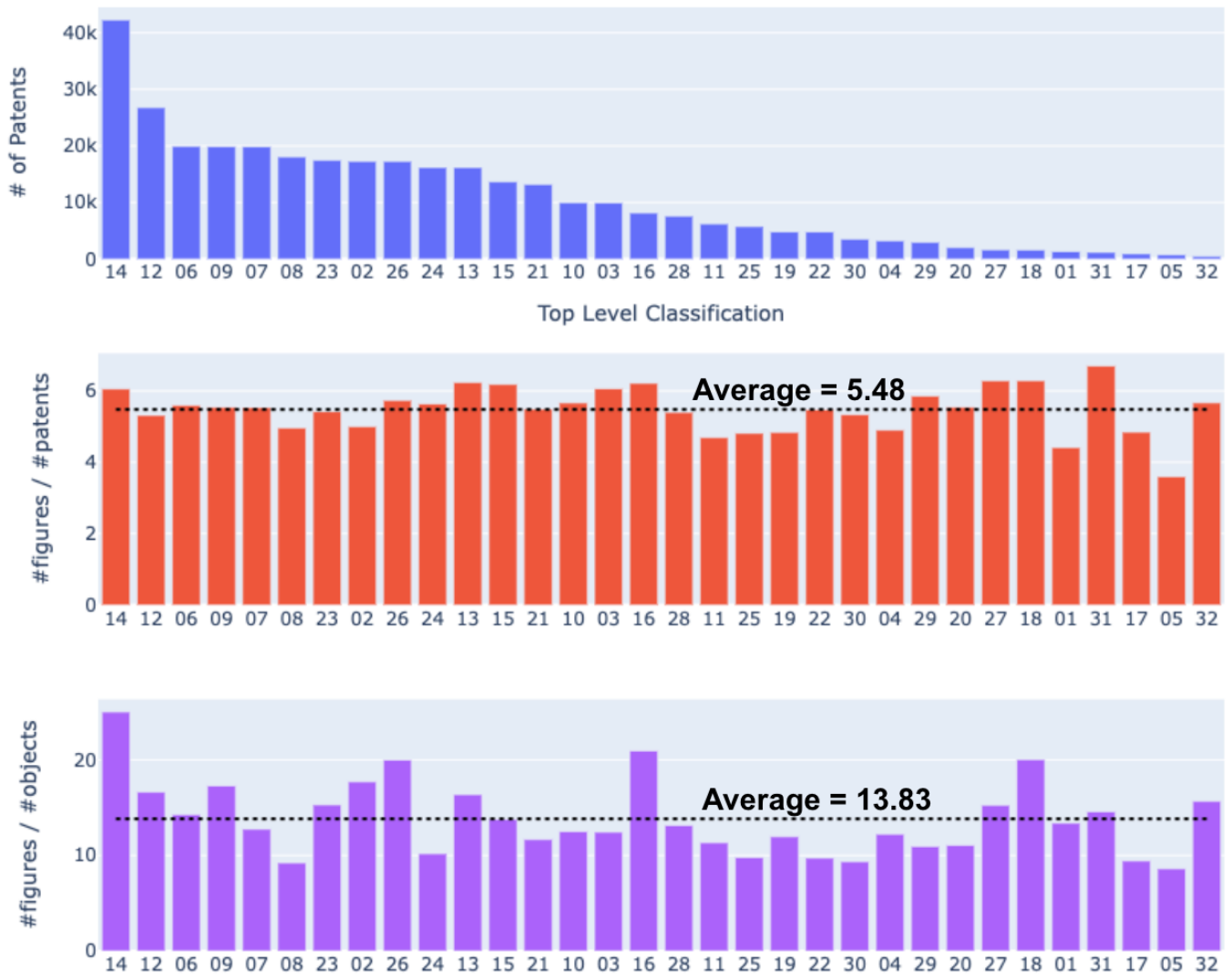}
    \caption{Distribution of patents, figures per patent, and figures per object identified over the patent classes. Only the top 32 classifications are shown. The remaining classifications are not listed because they contain less than 5 figures. The class codes in the abscissa correspond to the Locarno International Classification described in Table~\ref{tab:patentclassification}. }
    \label{fig:toplevelclassification}
\end{figure}
Table~\ref{tab:patentclassification} presents a list of top-level classes including the codes and descriptions. Figure~\ref{fig:toplevelclassification} illustrates the distribution of patents across different classes (top panel), the average number of figures per patent (middle panel), and the average number of figures per object (bottom panel) in DeepPatent2. The analysis reveals that, on average, each patent contains approximately 5.48 figures and each object is illustrated by on average 13.83 individual figures. 

\begin{figure}[H]
    \centering
    \includegraphics[width=0.5\textwidth]{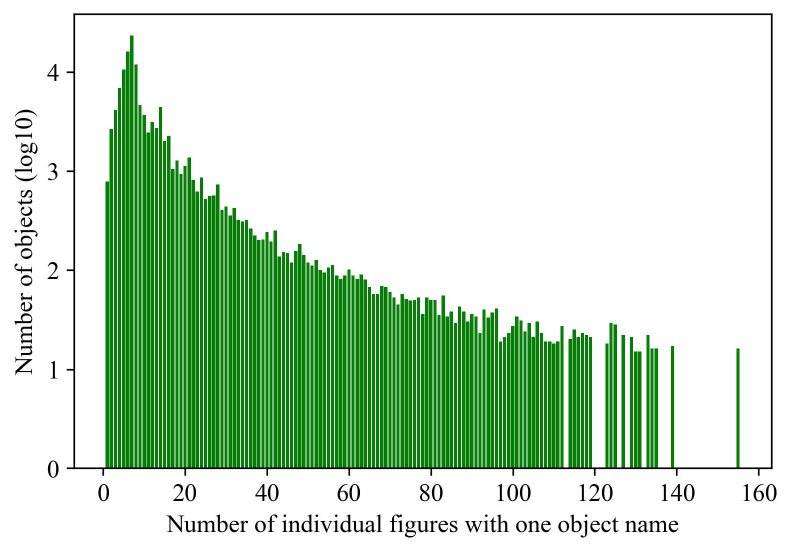}
    \caption{One object name may appear in multiple individual figures. This figure shows the distribution of the number of individual figures with one object name. For example, when the $x$-axis value is 7, the $y$-axis value is 23367, meaning each of the 23367 objects appears in 7 individual figures. The $x$-axis is truncated because data points beyond the truncation point are sparse.}
    \label{distribution_subfigure}
\end{figure}

The full dataset \cite{DVN/UG4SBD_2023} is publicly available under the Harvard Dataverse repository at 
\textcolor{blue}{ {\url{https://doi.org/10.7910/DVN/UG4SBD}}}.
For each year, the original and segmented figures are in separate tarball files, with a JSON file attached. 

\begin{table}[ht]
        \centering
        \begin{tabular}{c|l}
        \toprule
            \textbf{Class Code}& \textbf{Names} \\
            \midrule
            1 &  Foodstuffs\\
            2 &  Articles of clothing and haberdashery\\
            3 &  Travel goods, cases, parasols and personal belongings, not elsewhere specified\\
            4 & Brushware\\
            5 &  Textile piece goods, artificial and natural sheet material\\
            6 &  Furnishing\\
            7 &  Household goods, not elsewhere specified  \\
            8 &  Tools and hardware \\
            9 &  Packaging and containers for the transport or handling of goods \\
            10 &  Clocks and watches and other measuring instruments, checking and signaling instruments\\
            11 & Articles of adornment \\
            12 & Means of transport or hoisting \\
            13 & Equipment for production, distribution or transformation of electricity \\
            14 & Recording, telecommunication or data processing equipment \\
            15 & Machines, not elsewhere specified \\
            16 & Photographic, cinematographic, and optical apparatus \\
            17 & Musical instruments \\
            18 & Printing and office machinery \\
            19 & Stationery and office equipment, artists' and teaching materials \\
            20 & Sales and advertising equipment, signs \\
            21 & Games, toys, tents and sports goods \\
            22 & Arms, pyrotechnic articles, articles for hunting, fishing, and pest killing \\
            23 & Fluid distribution equipment, sanitary, heating, ventilation and air-conditioning equipment, solid fuel \\
            24 & Medical and laboratory equipment \\
            25 & Building units and construction elements \\
            26 & Lighting apparatus \\
            27 & Tobacco and smokers' supplies \\ 
            28 & Pharmaceutical and cosmetic products, toilet articles, and apparatus \\
            29 & Devices and equipment against fire hazards, for accident prevention and for rescue \\
            30 & Articles for the care and handling of animals \\
            31 & Machines and appliances for preparing food or drink, not elsewhere specified \\
            32 & Graphic symbols and logos, surface patterns, ornamentation, arrangement of interiors and exteriors \\

            \bottomrule
        \end{tabular}
        \caption{The top-level class codes and descriptions of the Locarno International Classification designation adopted by USPTO for design patents.}
        \label{tab:patentclassification}
\end{table}

\section*{Technical Validation}

\subsection*{Data Validation, Evaluation, and Copyright}\label{sec:validation}
Our data was automatically generated by high-performance machine learning and deep learning methods \cite{wei2022jcdl,hoque2022aaai} (read Section ``Methods" for a detailed description). However, errors can still occur, propagate, and accumulate leading to errors in the final data product. There are four possible error sources, namely, figure label detection, compound image segmentation, label association, and entity recognition (ER). Label association is dependent on image segmentation and OCR, but label association errors could occur even when image segmentation and OCR are correct.  
Assuming the label association (LA) can be approximated as the mismatch errors (Mismatch\% in Table~\ref{tab:count}), 
 
which is 7.5\% on average (so the precision $P_{\rm LA}=92.5\%$), the \emph{overall error rate} is calculated as 
\begin{equation}
\label{eq:error}
    E\approx1-P_{\rm LA}\times P_{\rm ER}=1-92.5\%\times96.0\%\approx 11.2\%.
\end{equation}
All figures are preserved in the final dataset, but mismatched figures are marked in their file names, so they can be used if needed. 
\begin{figure}[H]
    \centering
    \includegraphics[width=\textwidth]{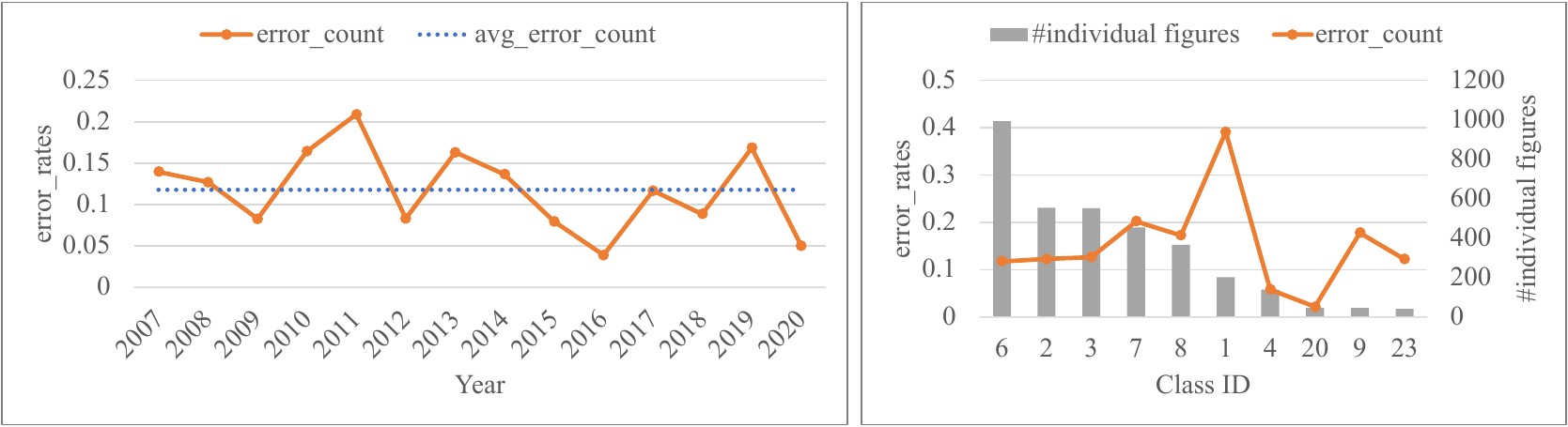}
    \caption{Left: Error rates calculated using the verification dataset over years; Right: Error rates across classes. The classes are ordered by the number of individual figures in each class. Class 6 (Furnishing) has the highest number of individual figures. Class 1 (Foodstuffs) has the highest error rate (0.39).}
    \label{fig:err}
\end{figure}
To validate the estimated error rates, we build a verification dataset by randomly sampling 100 compound figures each year from 2007 to 2020 for a total of 1400 compound figures. We manually inspected the quality of the final data product. Specifically, we inspected whether a compound figure is correctly segmented and whether the label, object name, and viewpoint of an individual figure are correctly extracted. We calculated the error rates by dividing the number of individual figures containing \emph{any} errors (404) by the total number of individual figures (3464). Figure~\ref{fig:err} (Left) shows the error rates calculated for each year and the average error rate. The verified error is consistent with the estimated error by Eq.~(\ref{eq:error}). Table~\ref{tab:errorrate} shows the breakdown of the verified errors by source. Note that the Verified Overall row is calculated using direct counts and should not be calculated from previous rows using Eq.(1), because a fraction of figures have more than one type of errors. We also calculate error rates for each class. The verification dataset contains 23 classes, of which 10 classes contain more than 40 individual figures. Figure~\ref{fig:err} (Right) shows the error rates across classes which have more than 40 individual figures in the verification dataset. 

Our estimated error rate is on par with other computer vision datasets, especially considering that our methods produce automated tags. Other auto-tagged computer vision datasets like WebVision estimate up to 20\% error rates \cite{li2017webvision}; while curated datasets like ImageNet-1k \cite{russakovsky2015imagenet1k} have been estimated to have error rates up to 6\% \cite{northcutt2021pervasive}. 

USPTO advocates open data, which can be freely used, reused, and redistributed \cite{usptoopendata}. The text and drawings of a patent are typically not subject to copyright restrictions \cite{uspto-tos}. Our dataset is under the Creative Commons Attribution (CC-BY 4.0) Generic International License.

\begin{table}[h]
    \centering
    \begin{tabular}{l|r}
    \toprule
    \multicolumn{1}{c|}{\bf Error Type}     &  \multicolumn{1}{c}{\bf Error}\\
   \midrule
   Estimated - Label (Segmentation+OCR) & 7.5\%\\
    Estimated - ER & 4\%\\
    \midrule
    \textbf{Estimated - Overall} & 11.2\% \\
    \midrule
    Verified - Segmentation & 5.52\% \\
    Verified - Label (excluding segmentation) & 1.87\% \\
    Verified - ER (Object extraction) & 7.61\% \\
    Verified - ER (Aspect extraction) & 0.91\%\\
    \midrule
    {\bf Verified - Overall} & 11.7\% \\
    \bottomrule
    \end{tabular}
    \caption{A breakdown analysis of the errors estimated using Eq.~(\ref{eq:error}) and the errors obtained using the verification dataset in the Technical Validation Section. The results show that they are consistent with a discrepancy of 0.5\%.}
    \label{tab:errorrate}
\end{table}

\section*{Usage Notes}

In this section, we demonstrate the usefulness and value of the dataset on a \emph{Conceptual Captioning} task, which generates short descriptive text that captures main objects and their viewpoints of a given image. 
This task was carried out on a dataset (hereafter CC18) consisting of natural images collected from the Web \cite{sharma2018conceptual}. In this work, we carry out this task on the technical drawings presented in our dataset.
We demonstrate how to use the dataset and advance the visual-to-text conversion task by running a state-of-the-art model on data in the technical drawing domain and potentially opening up research opportunities in other topics on patent drawings. In addition, we provide other tasks, in which our dataset could be potentially adopted. Here, we focus on demonstrating the potential benefit of improving the model performance by scaling up the training set using our dataset, instead of developing a new method.  

\subsection*{Conceptual Captioning}
The task of conceptual captioning involves generating a short textual description of an image. The state-of-the-art models usually employ the encoder-decoder architecture \cite{sharma2018conceptual}.  We employed ResNet-152 \cite{resnet}, a CNN-based model pre-trained on ImageNet \cite{russakovsky2015imagenet1k}, and fine-tuned on technical drawings selected from DeepPatent2. We varied the training data size from 500 to 1000 to 63,000 and tested the performance of each model on a corpus consisting of 600 figures with manually validated object names and viewpoints. The training sets were randomly selected from the parent sample. We selected the test set so it did not overlap with any samples from the training set. The input to the CNN encoder was an individual figure. The decoder generated a template-based caption including the object name and the viewpoint. The dimension of the encoder output was 2048 and we reduced it to 512 to match the input dimension of the LSTM decoder. We trained the image captioning network with the cross-entropy loss for 30 epochs with a mini-batch size of 32. We adopted an ADAM \cite{kingma2014adam} optimizer and a learning rate of 0.001. 

The models were evaluated using standard metrics including METEOR \cite{meteor}, NIST \cite{doddington2002nist}, Translation Error Rate (TER \cite{snover2006ter}), ROUGE \cite{rouge2004package}, and accuracy. The accuracy was defined as {a fraction of the ground-truth patent object names and viewpoints that were correctly predicted by the model}. The results in Table~\ref{tab:metrics} indicate that increasing the training size for automatically tagged data improved the performance of image captioning models.

The computing environment includes a Linux server with Intel Silver CPU, Nvidia GTX 2080 Ti. Because of the large size of data, we recommend a disk capacity of at least 500GB for all data. To obtain the subfigures from compound images, we used \emph{Pytorch >=1.4.0},  \emph{torchvision >= 0.5.0}, and \emph{Python 3.8}. The \emph{OpenCV} package was used to load PNG images. To load large JSON files, we recommend using \emph{ijson >3.1} package, which helps to conserve computer memory space. 

\begin{table}
\centering\small
 \begin{tabular}{ccccccccccc}
  \toprule
   {} & {} & {} & {} & {} & {} & {ROUGE-1} & {} & {} &{ROUGE-2} & {}\\
   \cmidrule{6-11}
    \textbf{Train Size} & {ACC.} & {METEOR} & {NIST} & {TER} & {P} & {R} & {F1} & {P} & {R} & {F1} \\
     \midrule
     \textbf{63000} & \textbf{0.647} & \textbf{0.627} & \textbf{1.62} & 0.49 & \textbf{0.645} & \textbf{0.641} & \textbf{0.634} & \textbf{0.545} & \textbf{0.544} & \textbf{0.534} \\  
 \textbf{1000} & 0.524 & 0.463 & 1.18 & 0.60 & 0.620 & 0.472 & 0.529 & 0.411 & 0.365 & 0.379 \\  
\textbf{500} & 0.518 & 0.451 & 1.11 & 0.60 & 0.620 & 0.447 & 0.513 & 0.405 & 0.343 & 0.365 \\  
  \bottomrule
  \end{tabular}
   \caption{A comparison of image captioning models with different training sizes. The best performance metrics are highlighted in bold.}
  \label{tab:metrics}
\end{table}

\subsection*{Potential Usages of the Dataset}
\begin{itemize}
    \item \textbf{Technical drawing image retrieval and semantic understanding.}  
    Investigation of effective and efficient retrieval methods for technical drawings has attracted the attention of the computer vision community. The Diagram and Abstract Imagery competition uses the DeepPatent dataset \cite{dirachallenge2022}. The rich semantic information of DeepPatent2 can potentially 
    help build a new multimodal (text+image) ground truth and enable tasks such as semantic understanding of abstract drawings and technical document classification \cite{jiang2021deeplearning}. 
    \item \textbf{Summarization of scholarly and technical corpora.} 
  
    Many existing summarization methods \cite{see_get_2017,MOIRANGTHEM20201,pegasus_zhang_2020} use only text. However, it has been shown that combining multimodal content improves a reader's understanding of each document's content while reducing the amount of content they must consume \cite{capra_augmenting_2013,jones_social_2019}. For example, the automatic selection of relevant images \cite{10.1145/3447535.3462505} has focused primarily on web resources and natural images. DeepPatent2, with its segmented compound figures, can be used to evaluate summary image selection in technical corpora.
   
    \item \textbf{3D image reconstruction.} Although humans are good at perceiving 3D objects from technical drawings, the task is challenging for computers. Delanoy et al.  developed neural methods to reconstruct 3D images from 2D sketches using training datasets corresponding to procedural, vases, and chairs \cite{delanoy20183d}. Our dataset contains  more diverse object types with multiple viewpoints and can potentially be used for training high-fidelity models. 
    \item \textbf{Figure segmentation.} Segmenting compound figures is a common preprocessing step before individual figures are used for analysis, retrieval, or machine learning tasks. Using 4000 DeepPatent2 figures, we achieved an accuracy of 99.5\% at both ${\rm IoU}=0.7$ and ${\rm IoU}=0.9$ evaluated on the technical drawings in \cite{hoque2022aaai}. Our data can potentially be used for training a large-scale base model that is fine-tunable for scientific or medical figure segmentation problems. 
    \item \textbf{Technical Drawing Classification.} Sketch classification methods \cite{zhang2016sketchnet, jearasuwan2019sketch,jiang2021deriving} have been proposed to recognize sketch images from the Web. However, these datasets contain a limited number of object categories and viewpoints. Our dataset contains more diverse object types and viewpoints and can potentially be used for training robust technical drawing classification models.
    \item \textbf{Create Generative and Multimodal Design Models for Innovation.} The Generative Adversarial Networks (GAN) and diffusion models perform remarkably well on text-to-image synthesis tasks \cite{dhariwal2021diffusion} and large language models (LLMs) have achieved superior performance on many generative NLP tasks \cite{zhao2023surveyllm}. Whether it is possible to combine LLMs with the diffusion models to automatically generate multimodal design models for innovation is an open question. Diffusion models and LLMs are known to be inaccurate in their details, and so the object name and viewpoint information of DeepPatent2 could provide the needed detailed technical drawings for training multimodal generative models that are accurate enough for design innovation. 
\end{itemize}

\section*{Code availability}

The code used for preprocessing and segmenting figures is publicly available on GitHub: \url{https://github.com/lamps-lab/Patent-figure-segmentor} and \url{https://github.com/GoFigure-LANL/figure-segmentation}. Similarly, the software used for extracting semantic information, including object names and viewpoints
from patent captions, is publicly available at \url{https://github.com/lamps-lab/Visual-Descriptor}.

\section*{Acknowledgments}
Research conducted by KA, XW, WS, MG, and JW presented in this paper was supported by Los Alamos National Laboratory subcontract BA601958 awarded to Old Dominion University. Research conducted by SJ, MK, and DO  was supported by the Laboratory Directed Research and Development program of Los Alamos National Laboratory under project number 20200041ER. 

\bibliographystyle{unsrt}
\bibliography{references}

\end{document}